\documentclass[nocompress]{spie}  

\usepackage{amsmath,amsfonts,amssymb}
\usepackage{graphicx}
\usepackage[colorlinks=true, allcolors=blue]{hyperref}
\usepackage{tabularx}
\usepackage{multirow}
\usepackage{multicol}
\usepackage{makecell}
\usepackage{booktabs}

\usepackage{orcidlink}
\usepackage[table]{xcolor}
\newcolumntype{Y}{>{\centering\arraybackslash}X}

\title{A Top-Down Framework for Metric-Scale Athlete Localization from Single Broadcast Frames}

\author[a,b]{Thanh-Khoi Nguyen}
\author[a,b]{Hoang-Phuc Nguyen}
\author[a,b]{Linh-Huynh}
\author[a,b]{Minh-Triet Tran $^{*}$}
\affil[a]{University of Science, VNU-HCM, Ho Chi Minh City, Vietnam}
\affil[b]{Vietnam National University, Ho Chi Minh City, Vietnam}

\makeatletter
\renewcommand{\cite}{\@ifnextchar[{\CITE@with}{\CITE@without}}
\def\CITE@without#1{[\citen{#1}]}
\def\CITE@with[#1]#2{[\citen{#2}, #1]}
\makeatother

\begin{document} 
\maketitle

\begin{abstract}
Accurate world-coordinate localization of athletes from single-frame broadcast footage is inherently challenging due to extreme scale disparities in ultra-high-resolution imagery. In this paper, we propose a top-down framework for metric-scale athlete localization from a single calibrated frame. Our approach centers on three key contributions. First, we propose \emph{Boundary-Aware Adaptive Tiling}, a semantics-guided extension of standard sliced inference. By iteratively expanding tile boundaries based on coarse bounding-box predictions, it systematically ensures full object containment, effectively mitigating boundary-splitting artifacts through a lightweight pipeline adaptation without architectural modifications. By substantially mitigating recall degradation under extreme scale variance, Boundary-Aware Adaptive Tiling enables us to isolate perspective distortion as the primary source of residual localization error. Second, we adapt the RTMPose-X architecture into a specialized two-keypoint estimator (pelvis and ground projection), employing a reformulated Gated Attention Unit optimized for this geometrically coupled point pair, and then deterministically lift the 2D ground projections into world coordinates via camera-calibrated ray casting. On the public test set, our method achieves a LocSim score of \textbf{97.44} and an mAP of \textbf{0.9128}, outperforming the baseline by over 21 \% and establishing a robust solution for high-resolution scale variance.
\end{abstract}

\section{Introduction}
\label{sec:intro}

Understanding the physical positions of athletes on a soccer pitch from camera imagery is a fundamental requirement for modern sports analytics~\cite{cioppa2021calibration,deepSoccerPositions}. Applications such as heatmap generation, ball-possession analysis, tactical understanding, and Game State Recognition (GSR) all rely on accurate metric-scale localization of every athlete visible in a scene~\cite{deepSoccerPositions,somers2024soccernet}. Motivated by this need, the recently introduced Spiideo SoccerNet SynLoc dataset~\cite{ardo2025synloc,cioppa2021calibration,splbev} formalizes the problem as a \emph{single-frame world-coordinate athlete detection and localization} task. In this setting, a model receives a 4K image captured by a static, calibrated camera covering half of a soccer pitch and is required to detect all athletes while estimating the 3D world coordinates (in meters) corresponding to the orthogonal projection of each athlete's pelvis onto the ground plane. Importantly, the evaluation metric, mAP-LocSim, is defined entirely in world space, emphasizing the importance of geometrically consistent localization rather than purely image-level representations~\cite{ardo2025synloc}.

Building upon this formulation, the primary objective of our work is to develop a robust top-down framework for metric-scale athlete localization from a single calibrated broadcast frame. The proposed system is designed to detect athletes across a wide range of apparent scales and accurately estimate the 3D world coordinates (in meters) of their orthogonal projections onto the pitch plane. To ensure a standardized and challenging evaluation protocol, we adopt the Spiideo SoccerNet SynLoc dataset as the benchmark for training and assessment of the proposed approach.

Two difficulties make this task non-trivial. First, \emph{detection at scale}: because a single camera covers half of a full-sized pitch, athletes near the far touchline occupy only a handful of pixels even in 4K imagery. Off-the-shelf person detectors trained on standard benchmarks fail to reliably localize such small targets when the full image is downsampled to a manageable inference resolution. Second, \emph{localization under perspective distortion}: mapping a 2D image observation to a metric pitch coordinate requires accurately identifying a specific anatomical landmark, the pelvis, rather than a coarse bounding-box surrogate such as the bottom-edge center, which the organizer baseline has shown to be a poor approximation of the true physical location~\cite{ardo2025synloc}. 
Standard multi-person pose estimators are ill-equipped for this strict geometric requirement. By dedicating capacity to modeling full-body skeletal dependencies, they over-parameterize the task, wasting computational resources on anatomically irrelevant joints. Consequently, they fail to exploit the camera's geometric properties, treating joints as independent visual targets rather than as points physically coupled by 3D projection.

We propose a two-stage top-down pipeline that addresses each bottleneck in turn. Rather than processing the full high-resolution frame at a reduced resolution, we first run a lightweight coarse detector to obtain approximate object locations, then use these as semantic constraints to guide tile boundary placement. This ensures every object is wholly enclosed within at least one crop at a consistent apparent scale, regardless of its distance from the camera. In the second stage, we observe that the two quantities required for world-coordinate localization - an anatomical reference point and its projection onto the ground plane - are not independent 2D targets but are physically coupled by the camera's perspective geometry: their relative displacement in the image encodes the subject's physical height under foreshortening. By strictly adapting the architecture to this physically coupled point pair, we propose a two-keypoint formulation that implicitly captures this geometric coupling, without auxiliary supervision or explicit camera-parameter input. Finally, the predicted 2D ground-projection keypoint is directly lifted to world coordinates via deterministic ray casting against the per-image camera calibration.

Extensive evaluations on the Spiideo SoccerNet SynLoc benchmark demonstrate the robustness of our approach. Our analysis reveals that our boundary-aware adaptive tiling detection strategy achieves the best performance across all metrics, isolating pose estimation as the primary challenge for metric-scale localization.
On the private test set, this configuration further achieves a LocSim score of 97.67, significantly outperforming the organizer baseline of 77.3. Furthermore, an ablation study in which we substitute predicted detections with ground-truth bounding boxes shows only a $\sim$1.4\% gap in LocSim, confirming that pose estimation, not detection, is the primary bottleneck for further improvement.

The main contributions of this work are as follows:
\begin{itemize}

    \item A \textbf{Geometrically Coupled Localization framework} that reformulates pose estimation into a specialized two-keypoint task (pelvis and ground projection). This allows the model to implicitly internalize perspective distortion and camera-projection physics within a 2D attention mechanism, thereby avoiding the over-parameterization of full-body pose estimators.
    

    \item \textbf{Boundary-Aware Adaptive Tiling (BAAT)}, a novel, semantics-guided cropping strategy. By using a coarse detection pass to iteratively expand tile boundaries, BAAT guarantees full object containment and eliminates boundary-splitting artifacts common in standard regular-grid sliced inference, without requiring additional annotations.
    
\end{itemize}

\section{Related Work}




\subsection{Small-Object Detection and Metric-Scale Athlete Localization}

Single-frame metric-scale athlete localization in high-resolution broadcast footage is inherently challenged by extreme scale variance and perspective distortion, as formalized by the SoccerNet SynLoc benchmark \cite{somers2024soccernet,ardo2025synloc}. Recovering these distant players is a specific instance of the broader small-object detection problem, canonically evaluated on datasets like VisDrone \cite{zhu2021visdrone}. While tiled inference methods such as Slicing Aided Hyper Inference (SAHI) improve small-object recall, their regular-grid approach introduces boundary-splitting artifacts, and existing adaptive variants often rely on geometric heuristics rather than semantic cues \cite{akyon2022sahi}. Although recent methods like ESOD \cite{liu2024esod} improve computational efficiency via feature-level slicing, they do not resolve boundary severing. Complementary to these efforts, our proposed Boundary-Aware Adaptive Tiling (BAAT) explicitly eliminates boundary-splitting by repurposing coarse semantic detections to iteratively expand tile boundaries, structurally guaranteeing full object containment.

\subsection{Geometry-Aware Pose Estimation Architectures}

Multi-person pose estimation is typically addressed via top-down or bottom-up paradigms. While bottom-up methods offer favorable scalability, top-down approaches decouple detection and keypoint estimation, enabling the modular integration of specialized components for higher accuracy. Recent top-down architectures, such as ViTPose~\cite{xu2022vitpose} and the RTMPose family~\cite{jiang2023rtmpose}, have substantially advanced state-of-the-art performance using transformer-based backbones and the SimCC one-dimensional coordinate classification representation~\cite{li2022simcc,jiang2023rtmpose}. However, these models are designed for full-body skeletons (e.g., 17 COCO joints), fundamentally over-parameterizing the SynLoc task, which evaluates performance exclusively in world space and requires only two geometrically coupled keypoints: the pelvis and its ground projection~\cite{ardo2025synloc}.

While the SynLoc baseline adopts this reduced two-keypoint representation, its reliance on a bottom-up formulation degrades both detection recall and projection accuracy compared to alternative paradigms. To address this, we adopt a top-down framework integrating a state-of-the-art YOLO26 detector~\cite{sapkota2025yolo26} with an adapted RTMPose-X model~\cite{jiang2023rtmpose}. By contracting the output space explicitly to the two-point formulation, our approach enables the network's Gated Attention Unit to focus exclusively on the physically coupled pelvis--ground-projection pair. This strictly constrained architecture allows the model to implicitly capture perspective foreshortening without requiring auxiliary geometric supervision.
\subsection{Sports Player Localization and Camera Calibration}

Localizing athletes in world coordinates requires tightly integrating visual detection with camera geometry. Prior works have addressed this within the SoccerNet framework by developing teacher-student calibration pipelines~\cite{cioppa2021calibration}, optimizing camera parameter estimation via keypoint and field-line correspondences~\cite{gutierrez2024pnlcalib}, and formalizing holistic Game State Reconstruction (GSR) for video sequences~\cite{somers2024gsr}. However, these approaches predominantly rely on temporal context or continuous tracking across frames. In contrast, our framework targets the strictly harder single-frame setting defined by the Spiideo SoccerNet SynLoc benchmark~\cite{ardo2025synloc}. By operating without temporal priors, our method must rely entirely on intra-frame geometric reasoning to achieve accurate metric-scale localization.

\section{Method}
\label{sec:method}

\begin{figure*}[h!]
    \centering
    \includegraphics[width=0.8\textwidth]{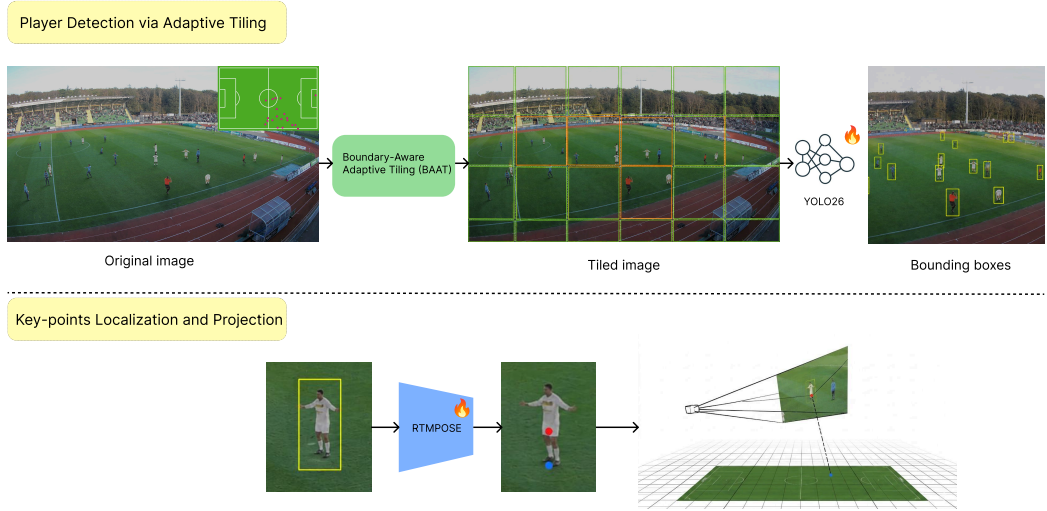}
    \caption{Overview of the proposed pipeline: boundary-aware tiled detection, two-keypoint pose estimation, and geometry-based world-coordinate projection.}
    \label{fig:overview}
\end{figure*}

We formulate the task as a two-stage top-down pipeline: (1) 
high-resolution player detection via adaptive tiling, (2) two-point 
keypoint estimation using a modified RTMPose-X model. The final 2D predictions are subsequently projected into world coordinates via deterministic ray casting. Our pipeline is demonstrated in Figure \ref{fig:overview}

\subsection{Stage 1: High-Resolution Player Detection via Adaptive Tiling}
\label{sec:detection}
To eliminate boundary-splitting artifacts inherent in standard sliced inference, we propose Boundary-Aware Adaptive Tiling (BAAT), a semantics-guided cropping strategy. First, a YOLO model processes the full 4K frame downsampled to $1280 \times 1280$ pixels to generate a set of coarse bounding boxes $\mathcal{D}=\{b_{i}\}_{i=1}^{N}$. Simultaneously, we initialize a standard SAHI grid $\mathcal{W}$ utilizing a $1280 \times 1280$ pixel slice size and a 0.2 spatial overlap.

Rather than relying purely on fixed geometry, BAAT utilizes the coarse detections $\mathcal{D}$ to iteratively expand the initial windows $\mathcal{W}$. For a given window $w=(x_{1},y_{1},x_{2},y_{2})$, let $\mathcal{B}(w) \subseteq \mathcal{D}$ denote the set of coarse boxes that intersect $w$ but are not fully contained within it. At each iteration, the window edges are expanded to enclose these intersecting boxes alongside a safety margin of $\delta=5$ pixels, clamped to the image width $W$ and height $H$:

\begin{equation}
\begin{alignedat}{2}
  x_1 &\leftarrow \max\!\left(0,\;
        \min_{b \in \mathcal{B}(w)} b_{x_1} - \delta\right), \qquad&
  y_1 &\leftarrow \max\!\left(0,\;
        \min_{b \in \mathcal{B}(w)} b_{y_1} - \delta\right), \\
  x_2 &\leftarrow \min\!\left(W,\;
        \max_{b \in \mathcal{B}(w)} b_{x_2} + \delta\right), \qquad&
  y_2 &\leftarrow \min\!\left(H,\;
        \max_{b \in \mathcal{B}(w)} b_{y_2} + \delta\right).
\end{alignedat}
\label{eq:adaptive-tiling-update}
\end{equation}

This expansion systematically ensures full object containment. The process repeats until no partially contained coarse box remains, bounded by a maximum of $T=100$ iterations, and converges empirically within three to five iterations.
\subsection{Stage 2: Two-Point Keypoint Estimation via Adapted RTMPose}
\label{sec:pose}
Standard top-down pose estimators are designed for full-body skeletons (e.g., 17 COCO joints), over-parameterizing our metric-localization task which requires only two geometrically coupled keypoints. To address this, we adapt the RTMPose-X~\cite{jiang2023rtmpose} architecture into a specialized two-point estimator by strictly contracting the output space of its RTMCCHead from $C_{\text{out}} = 17$ to $C_{\text{out}} = 2$, targeting exactly the pelvis and its orthogonal ground projection.

The RTMCCHead employs the SimCC one-dimensional coordinate classification~\cite{li2022simcc} and a Gated Attention Unit (GAU)~\cite{hua2022transformer} to refine keypoint representations. The GAU integrates a gated linear unit with self-attention, computing the output $O$ as:
$$A = \frac{1}{n}\mathrm{relu}^2\left(\frac{Q(X)K(Z)^\top}{\sqrt{s}}\right), \quad O = (U \odot AV)W_o$$
where $U, V$, and $Z$ are linear projections of the input $X$, $s$ is the scaling dimension, and $A$ is the attention matrix. By contracting the output dimension to $C_{\text{out}}=2$, the GAU sequence length $n$ naturally contracts to 2. Consequently, $A$ becomes a dense $2 \times 2$ attention matrix that exclusively models the relationship between the pelvis and the ground-projection keypoint. Because these two points are physically coupled by perspective geometry---their relative image displacement encodes the athlete's height under foreshortening---this contracted attention mechanism structurally incentivizes the network to implicitly internalize camera-projection physics without auxiliary supervision.

Finally, the predicted 2D ground-projection keypoint is deterministically mapped to metric world coordinates by un-projecting the point via the provided camera calibration and intersecting the resulting ray with the pitch plane ($Z=0$).

\section{Experiments}

\subsection{Experimental Setup}

\textbf{Dataset.} We evaluate on the Spiideo SoccerNet SynLoc benchmark~\cite{ardo2025synloc}, comprising 4K broadcast frames of synthetic athletes composited onto real-world backgrounds across 17 arenas. To rigorously test generalization, the public test (9.3K images) and private challenge (11.4K images) sets feature strictly held-out arenas unseen in the training split (42.5K images). The dataset provides image-space keypoints (pelvis and ground-projection), 3D metric ground truths, and explicit camera calibration parameters ($P=[R|t]$ and radial distortion polynomials) essential for world-coordinate ray casting.

\textbf{Metric and Baseline.} The primary evaluation metric is mAP-LocSim~\cite{ardo2025synloc}, which evaluates pure world-coordinate precision. It replaces bounding-box IoU with a metric-scale localization similarity function: 
\begin{equation}
    LocSim(d) = \exp\left(\frac{-\ln(0.05) \cdot d^2}{\tau^2}\right)
\end{equation}
where $d$ is the ground-plane Euclidean distance in meters between the predicted and true athlete position, and $\tau=1$m. We compare against the benchmark's baseline, a bottom-up YOLOX-pose framework that regresses the same two keypoints, achieving a LocSim of 76.17 on the public test set. Notably, standard 17-keypoint estimators cannot serve as valid baselines; canonical anatomical foot joints do not reliably lie on the pitch surface ($Z=0$) during dynamic actions, rendering them geometrically incompatible with deterministic ground-plane ray casting.

\subsection{Implementation Details}

\textbf{Detection:} We fine-tune YOLO26-Large~\cite{sapkota2025yolo26} on fixed SAHI $1280 \times 1280$ crops for 20 epochs. During inference, BAAT initializes with a $1280 \times 1280$ slice size and 0.2 spatial overlap. The coarse pass employs the same detector on the full 4K frame downsampled to $1280 \times 1280$, utilizing a deliberate 0.001 confidence threshold to maximize baseline recall prior to downstream pose estimation.

\textbf{Pose Estimation:} The adapted RTMPose-X~\cite{jiang2023rtmpose} is initialized from Body-7 pre-trained weights. Training utilizes the AdamW optimizer with a base learning rate of $10^{-4}$ and cosine annealing down to $5 \times 10^{-6}$ over 700 epochs (converging at epoch 62) with a batch size of 128 on a single GPU. Spatial augmentations include random horizontal flip, scaling, rotation, and CoarseDropout; Random HalfBody is explicitly excluded to prevent degenerate pelvis targets.

\subsection{Main Results}

\begin{figure}[htbp]
    \centering
    \includegraphics[width=0.45\linewidth]{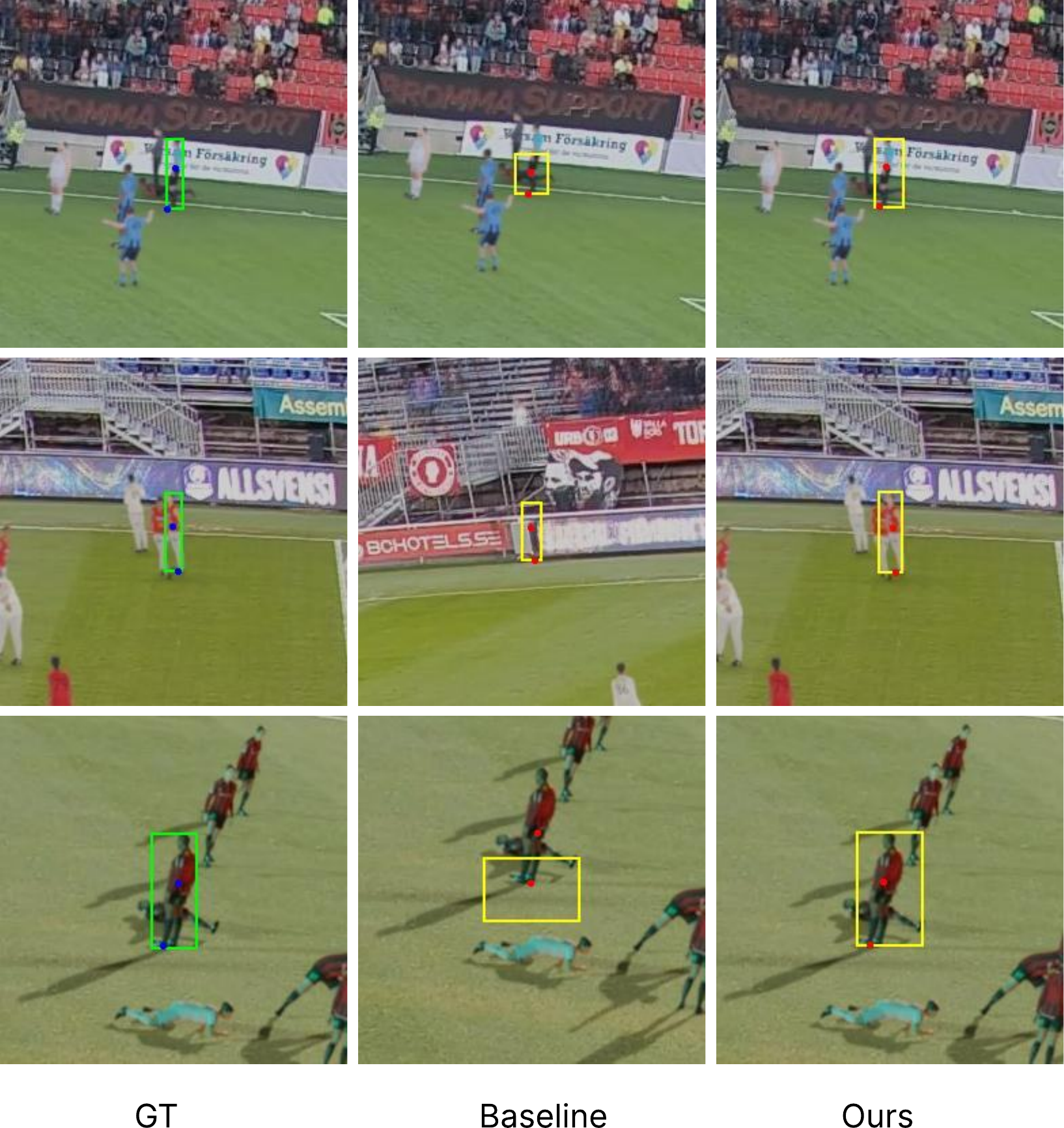}
    \caption{Qualitative comparison of athlete localization. Ours is more robust to scale variation and overlap, accurately predicting pelvis and ground-projection keypoints.}
    \label{fig:4_qualitative}
\end{figure}

As reported in Table 1, our framework significantly outperforms the organizer baseline across all metrics on the public test set, achieving a LocSim@$t=1$ of 97.44 (+21.27) and an AP@0.50:0.95 of 0.9128 (+21.70). This robust performance generalizes to the strictly held-out arenas of the private challenge set, attaining a LocSim of 96.67 compared to the baseline's 77.30. These substantial gains validate our sequential pipeline design: BAAT effectively saturates small-object detection recall across extreme scale disparities, isolating metric-scale precision under perspective distortion as the primary remaining error source. This distortion is subsequently mitigated by our geometrically coupled two-point pose formulation, demonstrating superior robustness over coarse bounding-box surrogates. Qualitative comparisons (Figure~\ref{fig:4_qualitative}) further confirm these capabilities, illustrating our method's ability to accurately recover distant athletes and cleanly resolve dense, overlapping cases where the baseline fails.
\begin{table}[t]
\centering
\caption{Comparison between the organizer baseline and the proposed method on the public-test and challenge sets. Green entries indicate improvements over the baseline. AP is unavailable for the challenge set.}
\label{tab:main-results}
\small
\setlength{\tabcolsep}{8pt}
\renewcommand{\arraystretch}{1.08}
\rowcolors{2}{gray!6}{white}
\begin{tabularx}{0.92\linewidth}{@{}l>{\raggedright\arraybackslash}Xccc@{}}
\toprule
\textbf{Set} & \textbf{Method} & \textbf{AP@0.50:0.95} & \textbf{LocSim@$t{=}1$} & \textbf{LocSim@$t{=}5$} \\
\midrule
Public test & Baseline & 0.6958 & 76.17 & 96.40 \\
Public test & Ours & \cellcolor{green!12}\textbf{0.9128} & \cellcolor{green!12}\textbf{97.44} & \cellcolor{green!12}\textbf{98.94} \\
Challenge & Baseline & -- & 77.30 & 96.63 \\
Challenge & Ours & -- & \cellcolor{green!12}\textbf{96.67} & \cellcolor{green!12}\textbf{98.98} \\
\bottomrule
\end{tabularx}
\end{table}

\section{Ablation study}
\paragraph{Detection variants.}
To understand the impact of the tiling strategy and detector training domain, we evaluate five configurations: \textbf{C} uses full-image training and full-image inference; \textbf{B1} uses SAHI-crop training and fixed SAHI inference; \textbf{B2} uses full-image-trained weights with fixed SAHI inference; \textbf{A1} uses SAHI-trained weights with Boundary-Aware Adaptive Tiling; and \textbf{A2} uses full-image-trained weights with Boundary-Aware Adaptive Tiling. In all five variants, the same adapted RTMPose-X pose estimator and ray-casting stage are applied downstream, so any performance differences are attributable solely to the detection stage.

\subsection{Effect of Tiling Strategy on Detection}

\begin{figure}[htbp]
    \centering
    \includegraphics[width=0.68\linewidth]{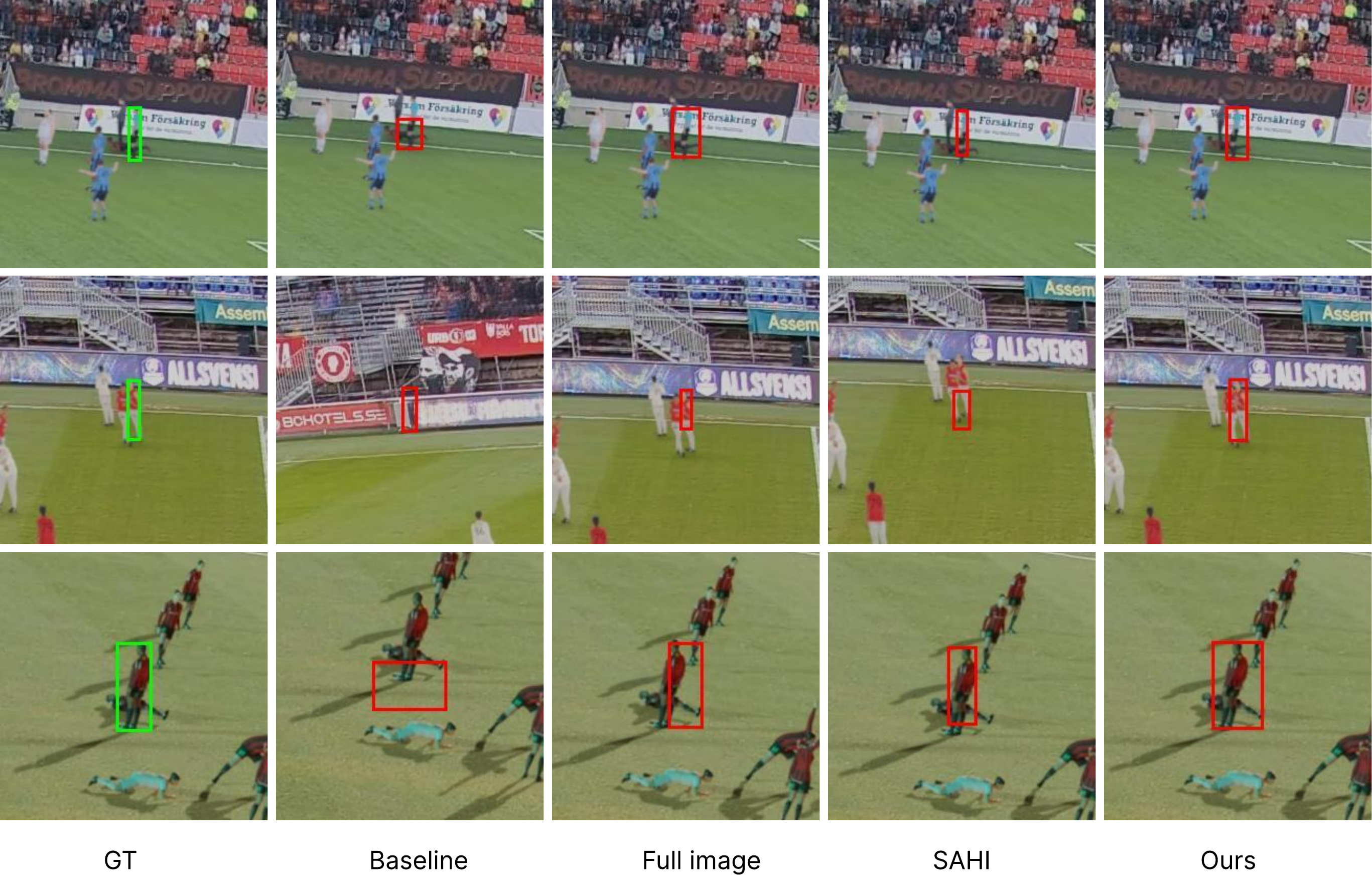}
    \caption{Qualitative detection outputs on a far-touchline crop under Approach~C (full-image), B2 (fixed SAHI), and A1 (Boundary-Aware Adaptive Tiling). Missed athletes are progressively recovered and boundary-splitting 
    artifacts are eliminated under A1.}
    \label{fig:5.1_detection_approaches_ablation}
\end{figure}

To isolate the impact of our tiling strategy and detector training domain, we evaluate five configurations (Table~\ref{tab:detection-ablation}). Full-image inference (Approach~C) establishes a weak baseline for small objects ($\text{AP}_\text{S}$ of 0.7668) due to resolution collapse at $1280\times1280$. While fixed SAHI tiling (Approach~B2) substantially recovers these missed instances ($\text{AP}_\text{S}$ of 0.8220), boundary-splitting artifacts impose a strict recall ceiling. Approach~A1, combining a SAHI-trained detector with Boundary-Aware Adaptive Tiling, eliminates these artifacts by construction, achieving the best overall performance (AP@0.50:0.95 of 0.9128, $\text{AR}_\text{S}$ of 0.8916). The marginal drop in $\text{AP}_\text{L}$ compared to Approach~B2 is an acceptable trade-off attributed to modest scale shifts during boundary expansion for large, nearby athletes. Finally, the consistent performance gap between Approaches A1 and A2 confirms that detectors must be trained on crops structurally aligned with their inference distribution. Qualitative comparisons are provided in Figure~\ref{fig:5.1_detection_approaches_ablation}.

\begin{table}[t]
\centering
\caption{Detection ablation on the public test set. Approach~A1 achieves the best performance across nearly all metrics. The exception is $\text{AP}_\text{L}$, where Approach~B2 marginally leads; see text for discussion. Best values per column are highlighted.}
\label{tab:detection-ablation}
\scriptsize
\setlength{\tabcolsep}{4.2pt}
\renewcommand{\arraystretch}{1.04}
\rowcolors{2}{gray!4}{white}
\resizebox{\linewidth}{!}{%
\begin{tabular}{@{}lcccccccccccc@{}}
\toprule
\textbf{Method} & \textbf{AP@0.50} & \textbf{AP@0.50:0.95} & \textbf{AP@0.75} 
& \textbf{AP$_\text{L}$} & \textbf{AP$_\text{M}$} & \textbf{AP$_\text{S}$} 
& \textbf{AR@1} & \textbf{AR@10} & \textbf{AR@100} 
& \textbf{AR$_\text{L}$} & \textbf{AR$_\text{M}$} & \textbf{AR$_\text{S}$} \\
\midrule
B1 & 0.9654 & 0.7895 & 0.8771 & 0.1808 & 0.8755 & 0.7521 
   & 0.0590 & 0.4451 & 0.8216 & 0.4232 & 0.9102 & 0.7664 \\
C  & 0.9764 & 0.8133 & 0.9002 & 0.3809 & 0.9033 & 0.7668 
   & 0.0595 & 0.4514 & 0.8382 & 0.5856 & 0.9285 & 0.7790 \\
B2 & 0.9792 & 0.8756 & 0.9363 & \cellcolor{green!12}\textbf{0.9743} & 0.9581 & 0.8220 
   & 0.0620 & 0.4791 & 0.8897 & 0.9927 & 0.9707 & 0.8306 \\
A2 & 0.9787 & 0.8811 & 0.9365 & 0.8862 & 0.9444 & 0.8496 
   & 0.0616 & 0.4825 & 0.9035 & 0.9872 & 0.9643 & 0.8590 \\
A1 & \cellcolor{green!12}\textbf{0.9895} & \cellcolor{green!12}\textbf{0.9128} 
   & \cellcolor{green!12}\textbf{0.9669} & 0.9728 
   & \cellcolor{green!12}\textbf{0.9652} & \cellcolor{green!12}\textbf{0.8822} 
   & \cellcolor{green!12}\textbf{0.0623} & \cellcolor{green!12}\textbf{0.4884} 
   & \cellcolor{green!12}\textbf{0.9286} & \cellcolor{green!12}\textbf{0.9955} 
   & \cellcolor{green!12}\textbf{0.9791} & \cellcolor{green!12}\textbf{0.8916} \\
\addlinespace[2pt]
\midrule
YOLOX-pose baseline & 0.9543 & 0.6958 & 0.7869 & 0.8020 & 0.8065 & 0.6277 
                    & 0.0556 & 0.4121 & 0.7420 & 0.9158 & 0.8699 & 0.6485 \\
\bottomrule
\end{tabular}}
\end{table}

\subsection{Implicit Geometric Learning in the Two-Keypoint Formulation}
\label{sec:geometric_consistency}

\begin{figure}[htbp]
    \centering
    \includegraphics[width=0.42\linewidth]{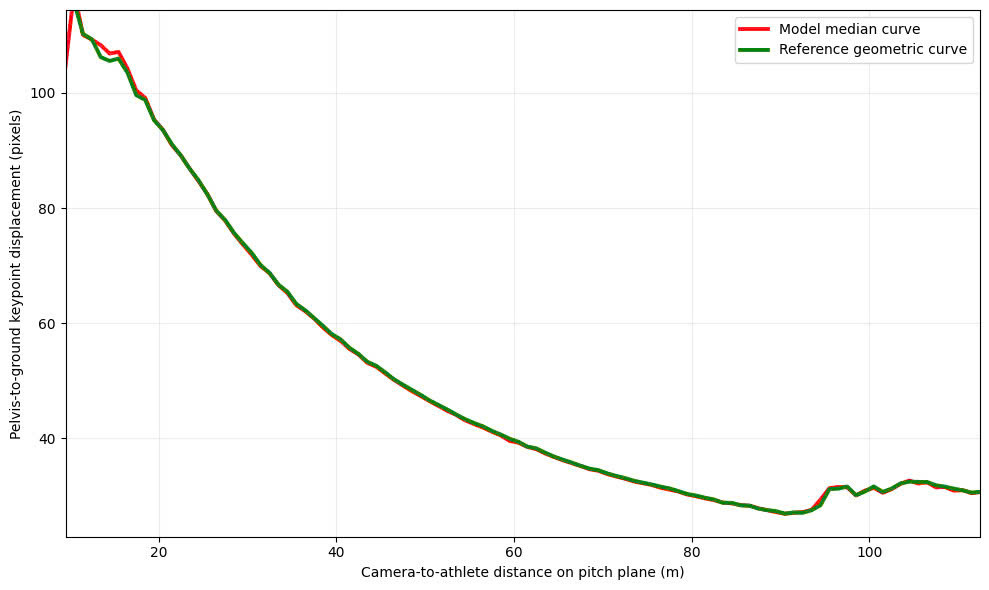}
    \caption{Predicted pelvis-to-ground displacement closely matches the reference geometric curve across camera-to-athlete distances (MAE 0.333px, corr. 0.8963). This confirms the two-keypoint model produces geometrically consistent outputs without camera parameter input.}
    \label{fig:geometry_ablation}
\end{figure}

To evaluate whether the two-keypoint formulation implicitly captures perspective geometry without receiving camera parameters as input, we compare the model's predicted pelvis-to-ground pixel displacement ($\Delta\hat{p} = \|\hat{p}_\text{pelvis} - \hat{p}_\text{ground}\|_2$) against a reference geometric curve. This physically correct reference is constructed by projecting 3D ground-truth annotations into image space using the provided camera calibration. Both sequences are then summarized as distance-binned medians and directly compared (Figure~\ref{fig:geometry_ablation}).

The predicted and reference curves align closely, following the same monotonic decrease expected from perspective foreshortening. As reported in Table~\ref{tab:geometry_ablation}, the predictions achieve a curve-level MAE of 0.333px and a Pearson correlation coefficient of 0.8963 across the pitch distance range. This high consistency establishes that restricting the model's capacity exclusively to this physically coupled point pair provides a structural incentive to preserve their geometric relationship in the predictions-a characteristic unlikely to emerge in full-body pose estimators burdened with anatomically irrelevant joints. The mean signed bias of -1.214px reflects a marginal tendency to underestimate displacement, predominantly localized to the sparsely sampled near-field range (7-12m, MAE = 24.830px). This data-driven bias identifies a concrete direction for future improvement through targeted data augmentation or geometry-aware loss functions.

\begin{table}[t]
\centering
\caption{Geometric consistency between predicted and reference displacement curves. Higher correlation and lower MAE indicate closer alignment with true perspective geometry.}
\label{tab:geometry_ablation}
\small
\setlength{\tabcolsep}{14pt}
\renewcommand{\arraystretch}{1.05}
\rowcolors{2}{gray!6}{white}
\begin{tabularx}{0.82\linewidth}{@{}>{\raggedright\arraybackslash}Xcc@{}}
\toprule
\textbf{Metric} & \textbf{Value} \\
\midrule
Curve-level MAE (px) $\downarrow$        & 0.333 \\
Correlation (pred vs.\ reference) $\uparrow$ & 0.8963 \\
\bottomrule
\end{tabularx}
\end{table}

\subsection{Detection Upper Bound}
To disentangle the contributions of the detection and pose estimation stages, we establish an upper bound by replacing the predicted bounding boxes from Approach~A1 with ground-truth annotations, holding the RTMPose-X estimator and ray-casting stages fixed. As reported in Table~\ref{tab:detection-upper-bound}, this perfect-detection scenario yields a LocSim of 98.86, representing a marginal absolute gain of 1.42 points over our predicted boxes (97.44). This narrow margin validates the effectiveness of Boundary-Aware Adaptive Tiling, confirming that the detection stage performs near-optimally and contributes minimally to residual localization error. Consequently, further improvements to the detection architecture or tiling strategy would yield diminishing returns. The primary bottleneck for end-to-end metric-scale localization lies squarely in the pose estimation stage-specifically, the spatial precision of the ground-projection keypoint under perspective distortion, occlusion, and extreme scale variance near the far touchline.

\begin{table}[t]
\centering
\caption{Detection upper-bound analysis on the public test set. Ground-truth bounding boxes replace predicted detections while all downstream stages remain fixed, isolating the contribution of the detection stage to the final LocSim.}
\label{tab:detection-upper-bound}
\small
\setlength{\tabcolsep}{10pt}
\renewcommand{\arraystretch}{1.05}
\rowcolors{2}{gray!6}{white}
\begin{tabular}{@{}lcc@{}}
\toprule
\textbf{Detection Input} & \textbf{LocSim} & \textbf{$\Delta$ vs. Predicted} \\
\midrule
Predicted Boxes (Approach A1) & 97.44 & --- \\
Ground-Truth Boxes (Upper Bound) & \textbf{98.86} & $+1.42$ $\uparrow$ \\
\bottomrule
\end{tabular}
\end{table}

\section{Conclusion}

We propose a two-stage top-down pipeline for metric-scale athlete localization from single-frame broadcast imagery. Boundary-Aware Adaptive Tiling eliminates boundary-splitting by adaptively expanding tiles using coarse detections, ensuring full object containment without annotations or architectural changes. A two-keypoint RTMPose-X formulation predicts pelvis and ground projection, enabling implicit perspective learning, with world coordinates recovered via ray casting. The method achieves 97.44 LocSim and 0.9128 AP, outperforming the baseline by over 21 points and generalizing to the private set.

Future work targets efficiency and geometric modeling. The sequential pipeline incurs latency due to per-instance pose inference, motivating batched inference, lightweight models, distillation, and hybrid designs. Incorporating geometry-aware supervision, such as calibration-based constraints, and leveraging monocular depth or geometric priors may improve robustness under occlusion, foreshortening, and scale variation, enhancing physical consistency.


\begingroup
\small
\bibliography{reference} 
\bibliographystyle{spiebib} 
\endgroup

\end{document}